%% file: main.tex
\documentclass{article}

\usepackage[preprint]{neurips_2021}

\usepackage[utf8]{inputenc} % allow utf-8 input
\usepackage[T1]{fontenc}    % use 8-bit T1 fonts
\usepackage{url}            % simple URL typesetting
\usepackage{booktabs}       % professional-quality tables
\usepackage{amsfonts}       % blackboard math symbols
\usepackage{nicefrac}       % compact symbols for 1/2, etc.
\usepackage{microtype}      % microtypography
\usepackage[dvipsnames]{xcolor}         % colors

\usepackage{amsmath,bm,amsthm,amssymb,mathtools}
\usepackage{thmtools,thm-restate}
\usepackage[scr=boondoxo,scrscaled=1.05]{mathalfa}
\input{math_commands.tex}

\usepackage{caption}
\usepackage{subcaption}
\usepackage{float}
\usepackage{graphicx}
\usepackage{multirow}
\usepackage{framed}
\usepackage{enumitem}
\usepackage{wrapfig}
\usepackage{algorithmic}
\usepackage{textcomp}
\usepackage{colortbl}
\usepackage{listings}
\usepackage{pifont}
\usepackage[most]{tcolorbox}
\usepackage{xstring}
\usepackage{xurl}
\usepackage{dsfont}
\usepackage{marvosym}
\usepackage{xspace}
\usepackage{epigraph}
\usepackage{gensymb}

\usepackage[
    pagebackref = true,
    colorlinks = true,
    linkcolor = MidnightBlue,
    urlcolor  = MidnightBlue,
    citecolor = MidnightBlue,
    anchorcolor = MidnightBlue
]{hyperref} 
\usepackage[nameinlink]{cleveref}
\renewcommand*\backref[1]{\ifx#1\relax \else (Cited on page #1) \fi}

\title{Transformers are Graph Neural Networks}

\author{%
  Chaitanya K. Joshi \\
  Department of Computer Science and Technology \\
  University of Cambridge, UK \\
  \href{mailto:chaitanya.joshi@cl.cam.ac.uk}{\texttt{chaitanya.joshi@cl.cam.ac.uk}}
}

\begin{document}

\maketitle

\begin{abstract}
  We establish connections between the Transformer architecture, originally introduced for natural language processing, and Graph Neural Networks (GNNs) for representation learning on graphs. 
  We show how Transformers can be viewed as message passing GNNs operating on fully connected graphs of tokens, where the self-attention mechanism capture the relative importance of all tokens w.r.t. each-other,
  and positional encodings provide hints about sequential ordering or structure.
  Thus, Transformers are expressive set processing networks that \emph{learn} relationships among input elements without being constrained by apriori graphs.
  Despite this mathematical connection to GNNs, Transformers are implemented via dense matrix operations that are significantly more efficient on modern hardware than sparse message passing.
  This leads to the perspective that Transformers are GNNs currently winning the hardware lottery.
\end{abstract}

\section{Transformers for Natural Language Processing}
\label{sec:nlp}

Our discussion focuses on \emph{representation learning}, which is the foundation for any machine learning task, be it predictive or generative modelling \citep{olah2014deep}.

At a high level, deep neural networks compress statistical and semantic information about data into a list of numbers, called a latent representation or embedding.
Models are trained by optimizing a loss function that measures how well the model's representations perform on a task of interest, such as predicting some properties of the input data.
For example, if we give a model a dataset of sentences and train it to predict the next word in each sentence, it will learn to build representations of each word that capture its meaning and context in the sentence \citep{graves2013generating}.

When trained on diverse but interconnected data sources, models learn to build general-purpose representations that capture the underlying structure of the data.
Good representations enable generalization to new tasks via knowledge transfer across related domains \citep{radford2019language, raffel2020exploring}. 
For instance, training a model on mathematical tasks may also improve it for programming related tasks, as both domains require abstract problem-solving.

A key ingredient for building good representations is a highly expressive and scalable model architecture. 
This can be understood by studying at the transformation of architectures in the context of Natural Language Processing (NLP).

%%%

\paragraph{From RNNs to Transformers}
Recurrent Neural Networks (RNNs) were a class of widely used deep learning architectures for NLP \citep{hochreiter1997long, sutskever2014sequence}.
RNNs build representations of each word in a sentence in a sequential manner, i.e. one word at a time. 
Intuitively, we can imagine an RNN layer as a conveyor belt, with the words being processed on it \textit{autoregressively} from left to right. 
At the end, we get a latent embedding for each word in the sentence, which we pass to the next RNN layer or use for our tasks of choice.

However, the sequential nature of RNNs means they struggle with long input contexts, as they compress the entire sentence into a single fixed-length representation at the end of the conveyor belt.
This lead to the design of the \emph{attention mechanism} \citep{bahdanau2014neural}, which allows RNNs to focus on different parts of the input sentence when building representations at each step, rather than just the representation of the last word processed.

Transformer networks \citep{vaswani2017attention}, which are built on top of the attention mechanism, take this idea further by allowing the model to build representations of each word in parallel, rather than sequentially.
This is done by computing the importance of each word in the sentence w.r.t. each other word, and then updating the representation of each word based on this importance.
This allows the model to capture long-range dependencies and relationships between words, leading to more expressive representations.

Initially introduced for machine translation\footnote{
  One of the motivations for Transformers was that it was inefficient to train RNNs on the large internal datasets that Google had for language translation. They needed to develop an expressive architecture that could scale better than RNNs.
}, Transformers have replaced RNNs as the architecture of choice across NLP \citep{achiam2023gpt} and wider deep learning applications \citep{dosovitskiy2020image, pmlr-v202-radford23a} due to their expressivity and scalability.

%%%

\begin{figure}[t!]
    \centering
    \includegraphics[width=0.8\linewidth]{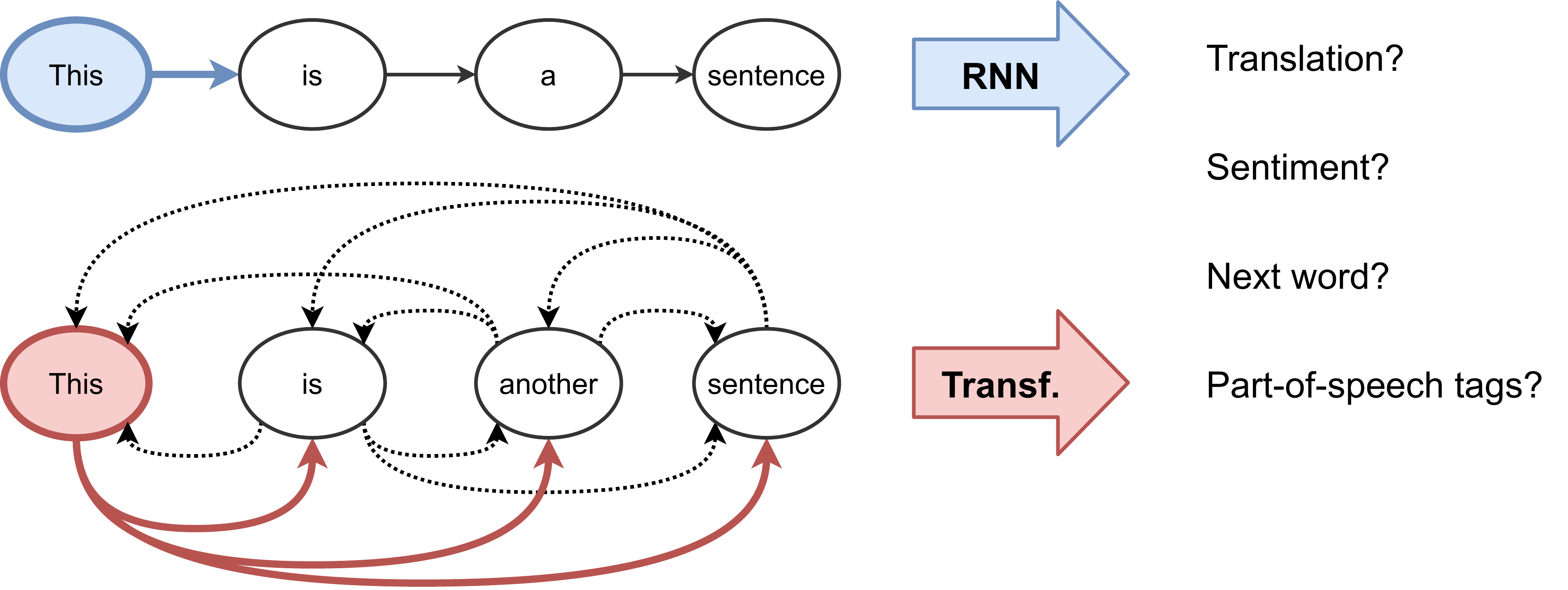}
    \caption{
    \textbf{Representation Learning for NLP.}
    RNNs build representations one token at a time, which captures the sequential nature of language.
    Transformers build representations in parallel via attention mechanisms, which capture relative importance of words w.r.t. each-other.
    }
    \label{fig:nlp}
\end{figure}

%%%

%%%

\paragraph{The attention mechanism}
The central component of the Transformer is the attention mechanism, which allows the representations of words in sentences to capture their relative importance w.r.t. each-other; see \citet{weng2018attention} for an intuitive introduction.

Formally, we are given a sentence $\mathcal{S}$ consisting of an ordered set of $n$ words (or tokens, more generally).
For each token $i$, we initialize its representation $h_i^{\ell=0} \in \mathbb{R}^{d}$ as an initial token embedding \citep{mikolov2013distributed}.
Token representations for each token $i$ are then updated via an attention mechanism from any arbitrary layer $\ell$ to layer $\ell+1$ as follows:
\begin{align}
  \label{eq:single-head-attention}
    h_{i}^{\ell+1} &= \text{Attention} \left( Q = W_Q^{\ell} \ h_{i}^{\ell}, \ K = \{ W_K^{\ell} \ h_{j}^{\ell} , \ \forall j \in \mathcal{S} \}, \ V = \{ W_V^{\ell} \ h_{j}^{\ell} , \ \forall j \in \mathcal{S} \} \right), \\
    &= \sum_{j \in \mathcal{S}} w_{ij} \cdot W_V^{\ell} \ h_{j}^{\ell} \ ,
\end{align}
where $j \in \mathcal{S}$ denotes the set of all tokens in the sentence (including token $i$ itself), and $W_Q^{\ell}, W_K^{\ell}, W_V^{\ell} \in \mathbb{R}^{d \times d}$ are learnable linear transformations denoting the \textbf{Q}uery, \textbf{K}ey and \textbf{V}alue for the attention computation, respectively.
The attention weights $w_{ij} \in \mathbb{R}$, which captures the relative importance between each pair of tokens $(i, j)$, are computed via a dot-product of the linearly transformed representations, followed by a softmax normalization across all tokens $j \in \mathcal{S}$:
\begin{align}
  \label{eq:softmax-attention}
    w_{ij} &= \text{softmax}_{j \in \mathcal{S}} \left( W_Q^{\ell} h_{i}^{\ell} \ \cdot \  W_K^{\ell} h_{j}^{\ell} \right), \\
    &= \frac{\text{exp} \left( W_Q^{\ell} h_{i}^{\ell} \ \cdot \ W_K^{\ell} h_{j}^{\ell} \right)}{\sum_{j' \in \mathcal{S}} \text{exp} \left( W_Q^{\ell} h_{i}^{\ell} \ \cdot \ W_K^{\ell} h_{j'}^{\ell} \right)} \ .
\end{align}

\begin{figure}[t!]
    \centering
    \includegraphics[width=0.65\linewidth]{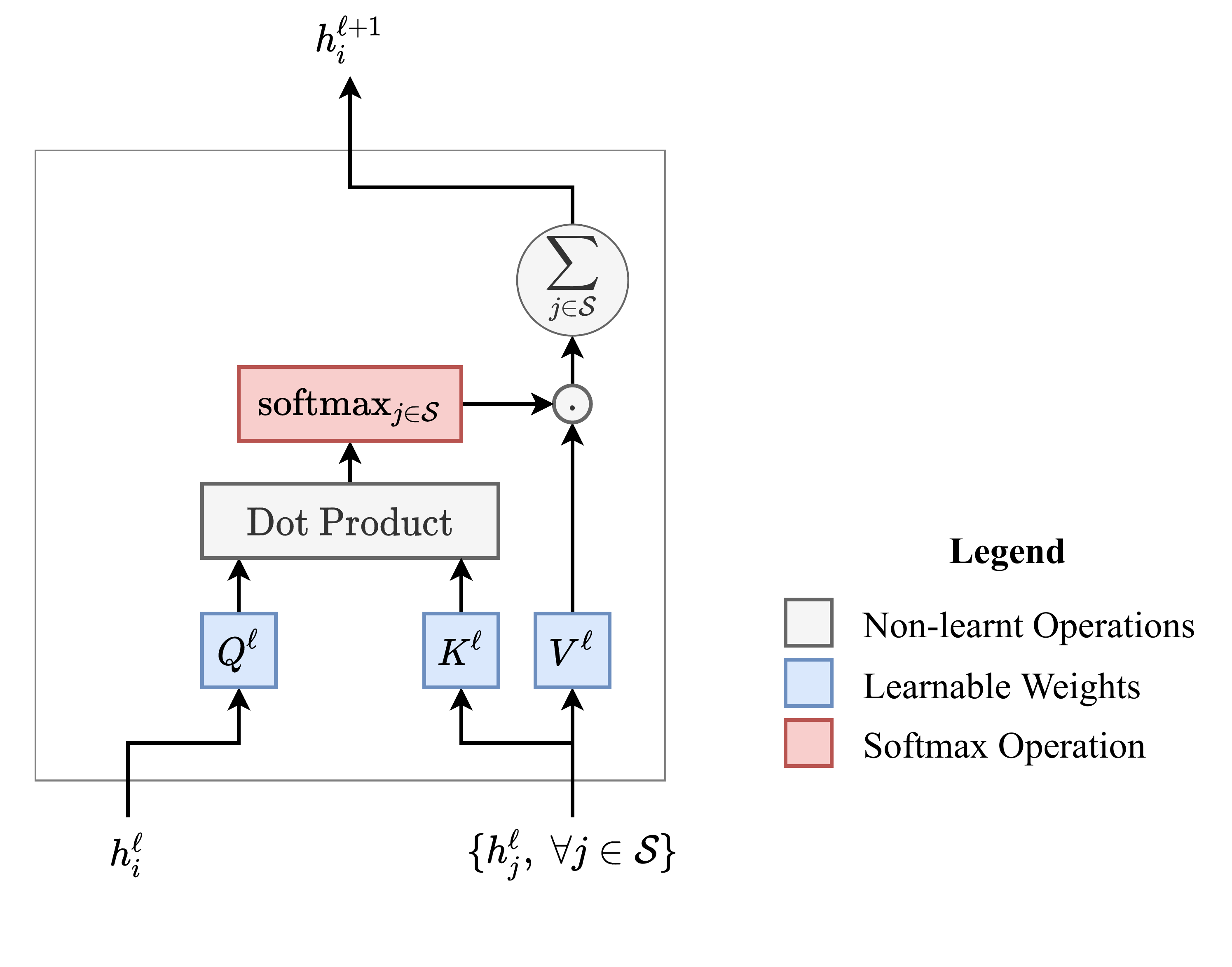}
    \caption{\textbf{A simple attention mechanism.} 
    Taking as input the representations of the token $h_{i}^{\ell}$ and the set of other tokens in the sentence $\{ h_{j}^{\ell} \;\ \forall j \in \mathcal{S} \}$, we compute the attention weights $w_{ij}$ denoting the relative importance for each pair $(i,j)$ through the dot-product followed by a softmax normalization. Finally, we produce the updated token representation $h_{i}^{\ell+1}$ by summing over the representations of tokens $\{ h_{j}^{\ell} \}$ weighted by the corresponding $w_{ij}$. 
    Each token in parallel undergoes the same pipeline to update its representation.
    }
    \label{fig:attention}
\end{figure}

\Cref{fig:attention} illustrates the described attention mechanism, which is a slightly simplified version of the one used in Transformers.
The attention computation in \eqref{eq:single-head-attention} is performed in parallel for each token in the sentence to obtain the updated representations in \textit{one shot}.
This is a key advantage of Transformers over RNNs, which update representations token-by-token.

\paragraph{Multi-head attention}

In practice, using a single attention head can be limiting, as it forces the model to learn only one set of weights that capture relationships among tokens.
However, tokens can interact in multiple ways, and we may want to simultaneously capture different aspects of their relationships, such as syntactic and semantic dependencies as well as contextual connections.

To improve the expressivity of the attention mechanism, Transformers introduced \emph{multi-head attention}, which computes multiple attention weights in parallel.
Each attention head learns its own set of query, key, and value transformations, allowing the model to attend to different types of relationships simultaneously and `hedge its bets' on the most relevant representations for each token.

Formally, we can define $K$ attention heads, where each head $k$ computes its own set of query, key, and value transformations when updating the representation of token $i$ at layer $\ell$:
\begin{align}
    Q^{k} = W_Q^{\ell, k} \ h_i^{\ell}, \quad
    K^{k} = \{ W_K^{\ell, k} \ h_j^{\ell} , \ \forall j \in \mathcal{S} \}, \quad
    V^{k,} = \{ W_V^{\ell, k} \ h_j^{\ell}, \ \forall j \in \mathcal{S} \},
\end{align}
where $W_Q^{\ell, k}, W_K^{\ell, k}, W_V^{\ell, k} \in \mathbb{R}^{d \times \frac{d}{k}}$ are learnable linear transformations for the $k^{\text{th}}$ attention head at layer $\ell$.
The output of each attention head is then computed as:
\begin{align}
    \text{head}_i^{k} &= \sum_{j \in \mathcal{S}} w_{ij}^{k} \cdot W_V^{\ell, k} \ h_{j}^{\ell} , \\
    \text{and} \quad w_{ij}^{k} &= \text{softmax}_{j \in \mathcal{S}} \left( W_Q^{\ell, k} \ h_{i}^{\ell} \ \cdot \ W_K^{\ell, k} \ h_{j}^{\ell} \right),
\end{align}
where $w_{ij}^{k} \in \mathbb{R}$ are the attention weights for the $k^{\text{th}}$ head, computed in the same way as before in \eqref{eq:softmax-attention}, but using the query and key transformations specific to that head.
The outputs of all attention heads are concatenated and projected to produce the updated representation for token $i$ as:
\begin{align}
    \label{eq:mhsa}
    \tilde h_{i}^{\ell} = \text{Concat} \left( \text{head}_i^1, \ldots, \text{head}_i^K \right) O^{\ell},
\end{align}
where $O^{\ell} \in \mathbb{R}^{d \times d}$ is a learnable linear transformation that projects the concatenated outputs to the original representation dimension $d$.
In practice, the computation for all the heads is done in parallel via batched matrix multiplication.

After the multi-head attention operation, each token undergoes an independent token-wise operation.
The updated representation $\tilde h_{i}^{\ell}$ for token $i$ is added to the original representation $h_i^{\ell}$ to form a residual connection \citep{he2016deep}, followed by a layer normalization \citep{ba2016layer}, and further processing by a token-wise multi-layer perceptron (MLP):
\begin{align}
    \label{eq:feedforward}
    h_i^{\ell+1} &= \text{MLP} \left( \text{LayerNorm} \left( h_i^{\ell} + \tilde h_i^{\ell} \right) \right),
\end{align}
where the MLP projects each token's representation to a higher dimension (typically $4 \times d$), applies a non-linear activation function, and then projects back to the original dimension $d$.
It is worth remarking that this token-wise MLP contains significantly more learnable parameters than the attention mechanism.

All of these operations (MLP sub-networks, residual connections, layer normalization) are now standard components and best practices of deep learning architectures.
They enable stacking multiple Transformer layers to build very deep networks, which has been key to the development of foundation models that learn complex relationships from large datasets \citep{Bommasani2021FoundationModels}.

\begin{figure}[t!]
    \centering
    \includegraphics[width=0.85\linewidth]{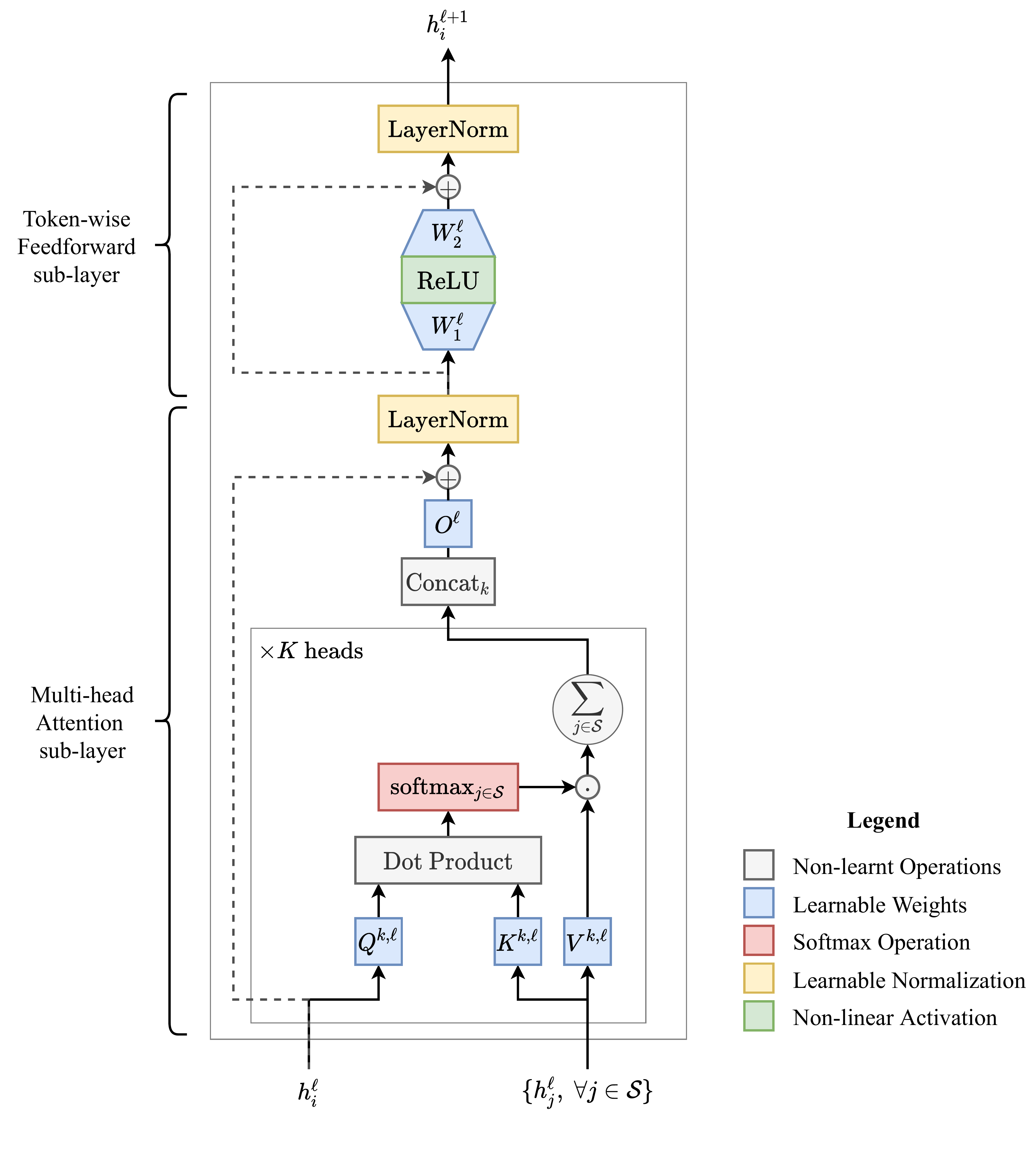}
    \caption{
      \textbf{A Transformer layer.}
      A multi-head attention sub-layer computes the relative importance of each token in a sentence w.r.t. each other token, and updates their representations accordingly.
      The updated representations are then processed by a token-wise multi-layer perceptron (MLP) sub-layer.
      Modern variants of the Transformer use SwiGLU \citep{shazeer2020glu} instead of ReLU \citep{glorot2011deep} as the MLP's activation function, and apply the LayerNorm operations before the multi-head attention and feed-forward sub-layers, rather than after \citep{xiong2020layer}.
    }
    \label{fig:transformer}
\end{figure}

The final picture of a Transformer layer is illustrated in \Cref{fig:transformer}.

%%%

\section{Graph Neural Networks for Representation Learning on Graphs}

Let us now turn our attention to sets with relational structure, i.e. graphs.

Graphs are used to model complex and interconnected systems in the real-world, ranging from knowledge graphs to social networks and molecular structures. 
Formally, an attributed graph $\mathcal{G} = ( \mA, \mH )$ is a set $\mathcal{V}$ of $n$ nodes connected by edges.
$\mA$ denotes an $n \times n$ adjacency matrix where each entry $a_{ij} \in \{ 0, 1 \}$ indicates the presence or absence of an edge connecting nodes $i$ and $j$.
Additionally, we can define $\mathcal{N}_i$ as the set of neighbors of node $i$, which are the nodes connected to $i$ by an edge, i.e. $\mathcal{N}_i = \{ j \in \mathcal{V} \mid a_{ij} = 1 \}$.
The matrix of initial representations $\mH \in \mathbb{R}^{n \times d}$ stores attributes $\vh_i \in \mathbb{R}^d$ associated with each node $i$.
For example, in molecular graphs, each node contains information about the atom type and edges can represent interactions among atoms.

Typically, the nodes in a graph have no canonical or fixed ordering and can be shuffled arbitrarily, resulting in an equivalent shuffling of the rows and columns of the adjacency matrix $\mA$.
Thus, accounting for permutation symmetry is a critical consideration when designing machine learning models for graphs \citep{bronstein2021geometric}.
Formally, a permutation $\sigma$ of the nodes acts on the graph $\mathcal{G}$ by permuting the rows and columns of the adjacency matrix $\mA$ and the representations $\mH$:
\begin{align}
  \mP_\sigma \gG &= (\mP_\sigma \mA \mP_\sigma^\top, \mP_\sigma \mH) ,
  \label{eq:perm}
\end{align}
where $\mP_\sigma$ is a permutation matrix corresponding to $\sigma$, and $\mP_\sigma \in \mathbb{R}^{n \times n}$ has exactly one 1 in every row and column, and 0 elsewhere.

\paragraph{Message Passing Graph Neural Networks}
Graph Neural Networks (GNNs) are a class of deep learning architectures designed to operate on graph-structured data.
GNNs leverage graph topology to propagate and aggregate information between connected nodes.
They have emerged as the architecture of choice for representation learning on graph data across domains ranging from molecular modelling \citep{stokes2020deep, batzner2022nequip} to recommendation systems \citep{ying2018graph} and transportation networks \citep{derrow2021eta}.

GNNs are based on the principle of \emph{message passing}, where each node iteratively updates its representations by aggregating from its local neighbors \citep{battaglia2018relational}. 
Formally, node representations $\vh_i$ for each node $i \in \mathcal{V}$ are updated from layer $\ell$ to $\ell+1$ through a three-step process: 
\begin{enumerate}
  \item \textit{Message construction}: For each node $i$ and its neighbors $j \in \mathcal{N}_i$, construct a message $\vm_{ij}^{\ell}$ that captures the relationship between the representations of nodes $i$ and $j$.
  \begin{align}
    \vm_{ij}^{\ell} &= \psi \left( \vh_i^{\ell}, \vh_j^{\ell} \right) , \quad \forall j \in \mathcal{N}_i,
    \label{eq:message}
  \end{align}
  where $\psi: \mathbb{R}^{2 \times d} \to \mathbb{R}^{d}$ is an MLP that learns to construct the message based on the representations of nodes $i$ and $j$.
  \item \textit{Aggregation}: Combine all messages from the neighbors of node $i$ to produce a single aggregated message $\vm_i^{\ell}$.
  \begin{align}
    \vm_i^{\ell} &= \bigoplus_{j \in \mathcal{N}_i} \vm_{ij}^{\ell} ,
    \label{eq:aggregate}
  \end{align}
  where $\bigoplus$ is a permutation-invariant operator (e.g. sum, mean, max) that aggregates messages from all neighbors $j \in \mathcal{N}_i$.
  Thus, a change in the order of neighbors does not affect the aggregated message, preserving the graph's symmetry (\eqref{eq:perm}).
  \item \textit{Update}: Update the representations of node $i$ using the aggregated message $\vm_i^{\ell}$ and its previous representations $\vh_i^{\ell}$.
  \begin{align}
    \vh_i^{\ell+1} &= \phi \left(\vh_i^{\ell} , \vm_i^{\ell} \right) ,
    \label{eq:update}
  \end{align}
  where $\phi: \mathbb{R}^{d} \to \mathbb{R}^{d}$ is another MLP.
\end{enumerate}

\begin{figure}[t!]
    \centering
    \includegraphics[width=\linewidth]{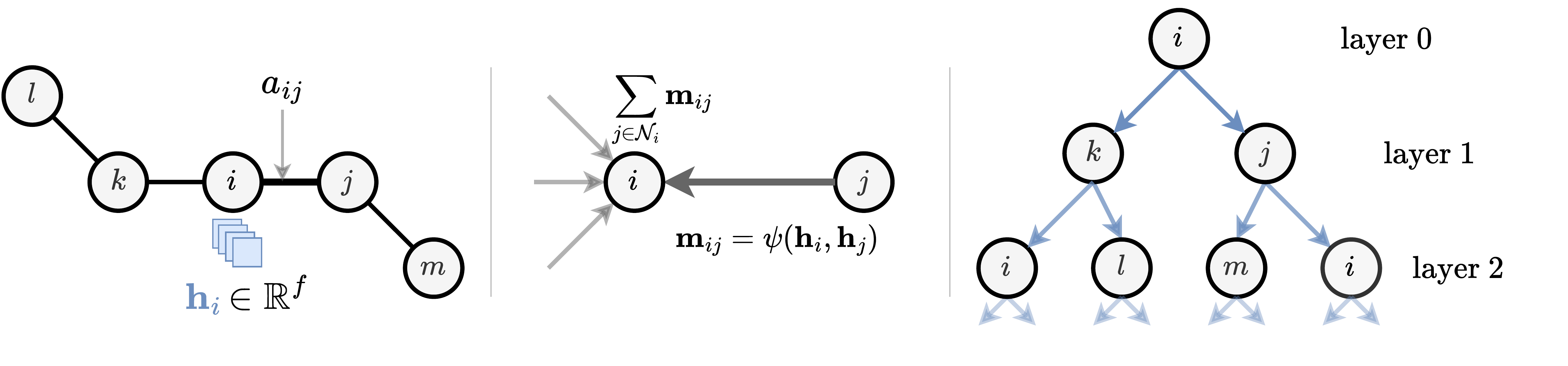}
    \caption{
    \textbf{Representation learning on graphs with message passing.}
    (left) Graphs model complex systems via a set of nodes connected by edges.
    (middle) GNNs build latent representations of graph data via \emph{message passing}, where each node learns to aggregate representations from its local neighbourhood.
    (right) Stacking $L$ message passing layers enables GNNs to send and aggregate information from $L$-hop subgraphs around each node.
    }
    \label{fig:gnn}
\end{figure}

This general formulation encompasses most commonly used GNN architectures including Graph Convolutional Networks \citep{kipf2017semi}, Graph Isomorphism Networks \citep{xu2018how}, and Message Passing Neural Networks \citep{gilmer2017neural}. 
By stacking multiple message passing layers, GNNs can propagate information beyond immediate neighbors and capture complex multi-hop relationships in the graph structure (\Cref{fig:gnn}).

\paragraph{Graph Attention Networks}
A particularly interesting class of GNNs uses attention mechanisms to weight the importance of different neighbors during aggregation \citep{velickovic2018graph}. 
In Graph Attention Networks (GATs), the message from neighbor $j$ to node $i$ is computed as:
\begin{align}
  \psi \left( \vh_i^{\ell}, \vh_j^{\ell} \right) &= \text{LocalAttention} \left( W^{\ell}_Q \ \vh_i^{\ell} \ , \ \{ W^{\ell}_K \ \vh_j^{\ell}, \ \forall j \in \mathcal{N}_i \} \ , \ \{ W^{\ell}_V \ \vh_j^{\ell}, \ \forall j \in \mathcal{N}_i \} \right) , \\
  &= \frac{\exp(W^{\ell}_Q \ \vh_i^{\ell} \cdot W^{\ell}_K \ \vh_j^{\ell})}{\sum_{j' \in \mathcal{N}_i} \exp(W^{\ell}_Q \ \vh_i^{\ell} \cdot W^{\ell}_K \ \vh_{j'}^{\ell})} \cdot W^{\ell}_V \ \vh_j^{\ell} \ ,
  \label{eq:gat_message}
\end{align}
where $W^{\ell}_Q, W^{\ell}_K, W^{\ell}_V \in \mathbb{R}^{d \times d}$.
The local attention mechanism allows allows GATs to learn which neighbors are more important for each node during the aggregation step.
The updated representation for node $i$ is computed by aggregating the messages from all its neighbors:
\begin{align}
  \vh_i^{\ell + 1} &= \vh_i^{\ell} \ + \ \sum_{j \in \mathcal{N}_i} \psi \left( \vh_i^{\ell}, \vh_j^{\ell} \right) ,
\end{align}
In practice, GATs also use multi-head attention to compute multiple sets of attention weights in parallel, allowing the model to learn different aspects of the relationships between nodes.

These equations should look very familiar!

In fact, they are almost identical to the Transformer's attention mechanism for computing the relative importance of words in a sentence.

%%%

\section{Transformers are GNNs over Fully Connected Graphs}

At this point, let us establish a formal equivalence between multi-head attention in Transformers and message passing in Graph Attention Networks (GATs).

Transformers can be viewed as GNNs operating on complete graphs, where self-attention models relationships between all input tokens in a sentence $\mathcal{S}$. 
The multi-head attention in \eqref{eq:mhsa} can be directly instantiated in the message passing framework as follows (for each head):
\begin{align}
  \psi \left( \vh_i^{\ell}, \vh_j^{\ell} \right) &= \text{GlobalAttention} \left( W^{\ell}_Q \ \vh_i^{\ell} \ , \ \{ W^{\ell}_K \ \vh_j^{\ell}, \ \forall j \in \mathcal{S} \} \ , \ \{ W^{\ell}_V \ \vh_j^{\ell}, \ \forall j \in \mathcal{S} \} \right) , \\
  &= \frac{\exp(W^{\ell}_Q \ \vh_i^{\ell} \cdot W^{\ell}_K \ \vh_j^{\ell})}{\sum_{j' \in \mathcal{S}} \exp(W^{\ell}_Q \ \vh_i^{\ell} \cdot W^{\ell}_K \ \vh_{j'}^{\ell})} \cdot W^{\ell}_V \ \vh_j^{\ell} \ .
\end{align}
Here, $\psi \left( \vh_i^{\ell} , \vh_j^{\ell} \right)$ computes the message from token $j$ to token $i$, with the relative importance of each token computed via attention.
Next, the weighted messages from all tokens in the sentence are aggregated via a summation, and the representation of token $i$ is updated as in \eqref{eq:feedforward} using a token-wise feedforward network $\phi$:
\begin{align}
\vh_i^{\ell+1} \ &= \ \phi \left( \vh_i^{\ell} \ , \ \vm_i^{\ell} \right) \
    = \text{MLP} \left( \text{LayerNorm} \left( h_i^{\ell} + \sum_{j \in \mathcal{S}} \psi \left( \vh_i^{\ell} , \vh_j^{\ell} \right) \right) \right) \ .
    \label{eq:trans_update}
\end{align}

We have arrived at exactly the same set of update equations as in \Cref{sec:nlp}.

Thus, Transformers are highly expressive set processing networks.
They can learn to capture both local and global context in the data via multi-head attention, without being constrained by the pathologies of pre-defined sparse graph structure as in GNNs \citep{di2023over}.
This is especially useful for tasks where we do not have an apriori graph structure, such as in modelling molecular structures. 
For instance, proteins fold into stable 3D structures through interactions between distant residues, so capturing global relationships is important for ensuring physical validity when predicting protein structures \citep{jumper2021highly}.

Conversely, GATs can be viewed as Transformers with attention restricted to local neighbourhoods, where the graph structure is used to implement sparse or masked attention \citep{dong2024flex}.

This connection has inspired a new development in graph representation learning.
Transformers use positional encodings as an initial feature to add information about the sequential ordering of tokens in a sentence, without strictly enforcing a sequential structure.
This idea can be extended to graphs, where positional encodings can be used to softly inject information about the graph structure into Transformer blocks \citep{dwivedi2021generalization}.
This has led to a new class of \emph{Graph Transformers} \citep{rampavsek2022recipe} that aim to combine both local message passing and global multi-head attention.
These architectures overcome the expressivity limitations of message passing GNNs while preserving the inductive bias of graph structure.

%%%

\section{Transformers are GNNs Winning the Hardware Lottery}

We will conclude with a discussion on the \emph{hardware lottery} \citep{hooker2021hardware}, the marriage of architectures and hardware that determines which research ideas rise to prominence, and its connection to the \emph{bitter lesson} in AI research \citep{sutton2019bitter}.

While we have established that Transformers are GNNs operating over fully connected graphs, there is an important practical difference between the two classes of architectures.

Transformers implement global multi-head attention through highly optimized \emph{dense} matrix multiplication operations that leverage the parallel processing capabilities of modern GPUs and TPUs.
Given a matrix of initial representations $\mH \in \mathbb{R}^{n \times d}$, the self-attention mechanism for all tokens in the sentence $\mathcal{S}$ can be computed in a parallel manner as follows:
\begin{align}
  \tilde \mH^{\ell} &= \text{softmax} \left( \mH^{\ell} \ W_Q^{\ell} \ \left( \mH^{\ell} \ W_K^{\ell} \right)^\top \right) \ \left( \mH^{\ell} \ W_V^{\ell} \right) ,
  \label{eq:transformer-matrix}
\end{align}
The multi-head attention mechanism can also be similarly parallelized by performing a single linear transformation for all heads in parallel, followed by reshaping the queries, keys, and values per head to tensors of dimension $n \times k \times d/k$ each.
The token-wise operations which follow multi-head attention are independent of other tokens and trivially parallelizable as well.

In contrast, GNNs typically perform \emph{sparse} message passing over locally connected structures, which is substantially less efficient on current GPUs for typical problem scales (with the exception of very sparse or billion-scale graphs).
This involves maintaining an index of neighbours for each node, and performing gather and scatter operations to propogate messages \citep{Fey/Lenssen/2019}.
As a result, GNNs are orders of magnitude slower to train and challenging to scale compared to standard Transformers on current hardware.

Additionally, with an increasing emphasis on scaling up models and datasets, there is empirical evidence that Transformers can learn the inductive biases baked in to GNNs, such as locality, when trained at sufficient scale.
This is particularly effective when models are given appropriate positional encodings as `hints' about the underlying structure of their input data, without imposing them as hard constraints in the architecture \citep{jaegle2021perceiver}.
The expressivity and flexibility of Transformers makes them the go-to architecture for representation learning on structured data across diverse application domains, including graphs.

Thus, Transformers are Graph Neural Networks that are currently winning the hardware lottery.

\paragraph{Acknowledgements}
I am grateful to numerous colleagues for feedback on this article over the years, which has helped shape my understanding of Transformers and GNNs.
I would also like to thank \href{https://thegradient.pub/}{The Gradient} for providing a platform to share an initial version of this work.

%%%

\newpage
\bibliographystyle{abbrvnat}
\bibliography{references}

\end{document}

%% file: math_commands.tex
\usepackage{amsmath,amsfonts,bm}
\usepackage{mathtools}

\def\eqref#1{equation~\ref{#1}}
\def\1{\bm{1}}

\def\vh{{\bm{h}}}

\def\vm{{\bm{m}}}

\def\mA{{\bm{A}}}

\def\mH{{\bm{H}}}

\def\mP{{\bm{P}}}

\DeclareMathAlphabet{\mathsfit}{\encodingdefault}{\sfdefault}{m}{sl}
\SetMathAlphabet{\mathsfit}{bold}{\encodingdefault}{\sfdefault}{bx}{n}

\def\gG{{\mathcal{G}}}

%% file: main.bbl
\begin{thebibliography}{39}
\providecommand{\natexlab}[1]{#1}
\providecommand{\url}[1]{\texttt{#1}}
\expandafter\ifx\csname urlstyle\endcsname\relax
  \providecommand{\doi}[1]{doi: #1}\else
  \providecommand{\doi}{doi: \begingroup \urlstyle{rm}\Url}\fi

\bibitem[Achiam et~al.(2023)Achiam, Adler, Agarwal, Ahmad, Akkaya, Aleman,
  Almeida, Altenschmidt, et~al.]{achiam2023gpt}
J.~Achiam, S.~Adler, S.~Agarwal, L.~Ahmad, I.~Akkaya, F.~L. Aleman, D.~Almeida,
  J.~Altenschmidt, et~al.
\newblock Gpt-4 technical report.
\newblock \emph{arXiv preprint arXiv:2303.08774}, 2023.

\bibitem[Ba et~al.(2016)Ba, Kiros, and Hinton]{ba2016layer}
J.~L. Ba, J.~R. Kiros, and G.~E. Hinton.
\newblock Layer normalization.
\newblock \emph{arXiv preprint arXiv:1607.06450}, 2016.

\bibitem[Bahdanau et~al.(2015)Bahdanau, Cho, and Bengio]{bahdanau2014neural}
D.~Bahdanau, K.~Cho, and Y.~Bengio.
\newblock Neural machine translation by jointly learning to align and
  translate.
\newblock In \emph{ICLR}, 2015.

\bibitem[Battaglia et~al.(2018)Battaglia, Hamrick, Bapst, Sanchez-Gonzalez,
  Zambaldi, et~al.]{battaglia2018relational}
P.~W. Battaglia, J.~B. Hamrick, V.~Bapst, A.~Sanchez-Gonzalez, V.~Zambaldi,
  et~al.
\newblock Relational inductive biases, deep learning, and graph networks.
\newblock \emph{arXiv preprint}, 2018.

\bibitem[Batzner et~al.(2022)Batzner, Musaelian, Sun, Geiger, Mailoa,
  Kornbluth, Molinari, Smidt, and Kozinsky]{batzner2022nequip}
S.~Batzner, A.~Musaelian, L.~Sun, M.~Geiger, J.~P. Mailoa, M.~Kornbluth,
  N.~Molinari, T.~E. Smidt, and B.~Kozinsky.
\newblock E (3)-equivariant graph neural networks for data-efficient and
  accurate interatomic potentials.
\newblock \emph{Nature communications}, 2022.

\bibitem[Bommasani et~al.(2021)Bommasani, Hudson, Adeli, Altman, Arora, von
  Arx, Bernstein, Bohg, Bosselut, Brunskill, Brynjolfsson,
  et~al.]{Bommasani2021FoundationModels}
R.~Bommasani, D.~A. Hudson, E.~Adeli, R.~Altman, S.~Arora, S.~von Arx, M.~S.
  Bernstein, J.~Bohg, A.~Bosselut, E.~Brunskill, E.~Brynjolfsson, et~al.
\newblock On the opportunities and risks of foundation models.
\newblock \emph{ArXiv}, 2021.

\bibitem[Bronstein et~al.(2021)Bronstein, Bruna, Cohen, and
  Veli{\v{c}}kovi{\'c}]{bronstein2021geometric}
M.~M. Bronstein, J.~Bruna, T.~Cohen, and P.~Veli{\v{c}}kovi{\'c}.
\newblock Geometric deep learning: Grids, groups, graphs, geodesics, and
  gauges.
\newblock \emph{arXiv preprint}, 2021.

\bibitem[Derrow-Pinion et~al.(2021)Derrow-Pinion, She, Wong, Lange, Hester,
  Perez, Nunkesser, Lee, Guo, Wiltshire, et~al.]{derrow2021eta}
A.~Derrow-Pinion, J.~She, D.~Wong, O.~Lange, T.~Hester, L.~Perez, M.~Nunkesser,
  S.~Lee, X.~Guo, B.~Wiltshire, et~al.
\newblock Eta prediction with graph neural networks in google maps.
\newblock In \emph{Proceedings of the 30th ACM international conference on
  information \& knowledge management}, 2021.

\bibitem[Di~Giovanni et~al.(2023)Di~Giovanni, Giusti, Barbero, Luise, Lio, and
  Bronstein]{di2023over}
F.~Di~Giovanni, L.~Giusti, F.~Barbero, G.~Luise, P.~Lio, and M.~M. Bronstein.
\newblock On over-squashing in message passing neural networks: The impact of
  width, depth, and topology.
\newblock In \emph{International Conference on Machine Learning}. PMLR, 2023.

\bibitem[Dong et~al.(2024)Dong, Feng, Guessous, Liang, and He]{dong2024flex}
J.~Dong, B.~Feng, D.~Guessous, Y.~Liang, and H.~He.
\newblock Flex attention: A programming model for generating optimized
  attention kernels.
\newblock \emph{arXiv preprint arXiv:2412.05496}, 2024.

\bibitem[Dosovitskiy et~al.(2021)Dosovitskiy, Beyer, Kolesnikov, Weissenborn,
  Zhai, Unterthiner, Dehghani, Minderer, Heigold, Gelly,
  et~al.]{dosovitskiy2020image}
A.~Dosovitskiy, L.~Beyer, A.~Kolesnikov, D.~Weissenborn, X.~Zhai,
  T.~Unterthiner, M.~Dehghani, M.~Minderer, G.~Heigold, S.~Gelly, et~al.
\newblock An image is worth 16x16 words: Transformers for image recognition at
  scale.
\newblock In \emph{International Conference on Learning Representations,
  {ICLR}}, 2021.

\bibitem[Dwivedi and Bresson(2020)]{dwivedi2021generalization}
V.~P. Dwivedi and X.~Bresson.
\newblock A generalization of transformer networks to graphs.
\newblock \emph{arXiv preprint arXiv:2012.09699}, 2020.

\bibitem[Fey and Lenssen(2019)]{Fey/Lenssen/2019}
M.~Fey and J.~E. Lenssen.
\newblock Fast graph representation learning with {PyTorch Geometric}.
\newblock In \emph{ICLR Workshop on Representation Learning on Graphs and
  Manifolds}, 2019.

\bibitem[Gilmer et~al.(2017)Gilmer, Schoenholz, Riley, Vinyals, and
  Dahl]{gilmer2017neural}
J.~Gilmer, S.~S. Schoenholz, P.~F. Riley, O.~Vinyals, and G.~E. Dahl.
\newblock Neural message passing for quantum chemistry.
\newblock In \emph{ICML}, 2017.

\bibitem[Glorot et~al.(2011)Glorot, Bordes, and Bengio]{glorot2011deep}
X.~Glorot, A.~Bordes, and Y.~Bengio.
\newblock Deep sparse rectifier neural networks.
\newblock In \emph{Proceedings of the fourteenth international conference on
  artificial intelligence and statistics}, 2011.

\bibitem[Graves(2013)]{graves2013generating}
A.~Graves.
\newblock Generating sequences with recurrent neural networks.
\newblock \emph{arXiv preprint arXiv:1308.0850}, 2013.

\bibitem[He et~al.(2016)He, Zhang, Ren, and Sun]{he2016deep}
K.~He, X.~Zhang, S.~Ren, and J.~Sun.
\newblock Deep residual learning for image recognition.
\newblock In \emph{Proceedings of the IEEE conference on computer vision and
  pattern recognition}, pages 770--778, 2016.

\bibitem[Hochreiter and Schmidhuber(1997)]{hochreiter1997long}
S.~Hochreiter and J.~Schmidhuber.
\newblock Long short-term memory.
\newblock \emph{Neural computation}, 1997.

\bibitem[Hooker(2021)]{hooker2021hardware}
S.~Hooker.
\newblock The hardware lottery.
\newblock \emph{Communications of the ACM}, 2021.

\bibitem[Jaegle et~al.(2021)Jaegle, Borgeaud, Alayrac, Doersch, Ionescu, Ding,
  Koppula, Zoran, Brock, Shelhamer, et~al.]{jaegle2021perceiver}
A.~Jaegle, S.~Borgeaud, J.-B. Alayrac, C.~Doersch, C.~Ionescu, D.~Ding,
  S.~Koppula, D.~Zoran, A.~Brock, E.~Shelhamer, et~al.
\newblock Perceiver io: A general architecture for structured inputs \&
  outputs.
\newblock \emph{arXiv preprint arXiv:2107.14795}, 2021.

\bibitem[Joshi(2020)]{joshi2020transformers}
C.~Joshi.
\newblock Transformers are graph neural networks.
\newblock \emph{The Gradient}, 2020.
\newblock URL
  \url{https://thegradient.pub/transformers-are-gaph-neural-networks/}.

\bibitem[Jumper et~al.(2021)Jumper, Evans, Pritzel, Green, Figurnov,
  Ronneberger, et~al.]{jumper2021highly}
J.~Jumper, R.~Evans, A.~Pritzel, T.~Green, M.~Figurnov, O.~Ronneberger, et~al.
\newblock Highly accurate protein structure prediction with alphafold.
\newblock \emph{Nature}, 2021.

\bibitem[Kipf and Welling(2017)]{kipf2017semi}
T.~N. Kipf and M.~Welling.
\newblock Semi-supervised classification with graph convolutional networks.
\newblock In \emph{ICLR}, 2017.

\bibitem[Mikolov et~al.(2013)Mikolov, Sutskever, Chen, Corrado, and
  Dean]{mikolov2013distributed}
T.~Mikolov, I.~Sutskever, K.~Chen, G.~S. Corrado, and J.~Dean.
\newblock Distributed representations of words and phrases and their
  compositionality.
\newblock \emph{Advances in neural information processing systems}, 2013.

\bibitem[Olah(2014)]{olah2014deep}
C.~Olah.
\newblock Deep learning, {NLP}, and representations.
\newblock \emph{Christopher Olah's Blog}, 2014.
\newblock URL
  \url{https://colah.github.io/posts/2014-07-NLP-RNNs-Representations/}.

\bibitem[Radford et~al.(2019)Radford, Wu, Child, Luan, Amodei, and
  Sutskever]{radford2019language}
A.~Radford, J.~Wu, R.~Child, D.~Luan, D.~Amodei, and I.~Sutskever.
\newblock Language models are unsupervised multitask learners.
\newblock \emph{OpenAI}, 2019.

\bibitem[Radford et~al.(2023)Radford, Kim, Xu, Brockman, Mcleavey, and
  Sutskever]{pmlr-v202-radford23a}
A.~Radford, J.~W. Kim, T.~Xu, G.~Brockman, C.~Mcleavey, and I.~Sutskever.
\newblock Robust speech recognition via large-scale weak supervision.
\newblock In \emph{International Conference on Machine Learning}, 2023.

\bibitem[Raffel et~al.(2020)Raffel, Shazeer, Roberts, Lee, Narang, Matena,
  Zhou, Li, and Liu]{raffel2020exploring}
C.~Raffel, N.~Shazeer, A.~Roberts, K.~Lee, S.~Narang, M.~Matena, Y.~Zhou,
  W.~Li, and P.~J. Liu.
\newblock Exploring the limits of transfer learning with a unified text-to-text
  transformer.
\newblock \emph{Journal of machine learning research}, 2020.

\bibitem[Ramp{\'a}{\v{s}}ek et~al.(2022)Ramp{\'a}{\v{s}}ek, Galkin, Dwivedi,
  Luu, Wolf, and Beaini]{rampavsek2022recipe}
L.~Ramp{\'a}{\v{s}}ek, M.~Galkin, V.~P. Dwivedi, A.~T. Luu, G.~Wolf, and
  D.~Beaini.
\newblock Recipe for a general, powerful, scalable graph transformer.
\newblock \emph{Advances in Neural Information Processing Systems}, 2022.

\bibitem[Shazeer(2020)]{shazeer2020glu}
N.~Shazeer.
\newblock Glu variants improve transformer.
\newblock \emph{arXiv preprint arXiv:2002.05202}, 2020.

\bibitem[Stokes et~al.(2020)Stokes, Yang, Swanson, Jin, Cubillos-Ruiz, Donghia,
  MacNair, French, Carfrae, Bloom-Ackermann, et~al.]{stokes2020deep}
J.~M. Stokes, K.~Yang, K.~Swanson, W.~Jin, A.~Cubillos-Ruiz, N.~M. Donghia,
  C.~R. MacNair, S.~French, L.~A. Carfrae, Z.~Bloom-Ackermann, et~al.
\newblock A deep learning approach to antibiotic discovery.
\newblock \emph{Cell}, 2020.

\bibitem[Sutskever et~al.(2014)Sutskever, Vinyals, and
  Le]{sutskever2014sequence}
I.~Sutskever, O.~Vinyals, and Q.~V. Le.
\newblock Sequence to sequence learning with neural networks.
\newblock In \emph{Advances in neural information processing systems}, 2014.

\bibitem[Sutton(2019)]{sutton2019bitter}
R.~Sutton.
\newblock The bitter lesson.
\newblock \emph{Incomplete Ideas (blog)}, 2019.

\bibitem[Vaswani et~al.(2017)Vaswani, Shazeer, Parmar, Uszkoreit, Jones, Gomez,
  Kaiser, and Polosukhin]{vaswani2017attention}
A.~Vaswani, N.~Shazeer, N.~Parmar, J.~Uszkoreit, L.~Jones, A.~N. Gomez,
  L.~Kaiser, and I.~Polosukhin.
\newblock Attention is all you need.
\newblock In \emph{Advances in neural information processing systems}, 2017.

\bibitem[Veli{\v{c}}kovi{\'{c}} et~al.(2018)Veli{\v{c}}kovi{\'{c}}, Cucurull,
  Casanova, Romero, Li{\`{o}}, and Bengio]{velickovic2018graph}
P.~Veli{\v{c}}kovi{\'{c}}, G.~Cucurull, A.~Casanova, A.~Romero, P.~Li{\`{o}},
  and Y.~Bengio.
\newblock {Graph Attention Networks}.
\newblock \emph{ICLR}, 2018.

\bibitem[Weng(2018)]{weng2018attention}
L.~Weng.
\newblock Attention? attention!
\newblock \emph{Lil'Log}, 2018.
\newblock URL
  \url{http://lilianweng.github.io/lil-log/2018/06/24/attention-attention.html}.

\bibitem[Xiong et~al.(2020)Xiong, Yang, He, Zheng, Zheng, Xing, Zhang, Lan,
  Wang, and Liu]{xiong2020layer}
R.~Xiong, Y.~Yang, D.~He, K.~Zheng, S.~Zheng, C.~Xing, H.~Zhang, Y.~Lan,
  L.~Wang, and T.~Liu.
\newblock On layer normalization in the transformer architecture.
\newblock In \emph{International conference on machine learning}, 2020.

\bibitem[Xu et~al.(2019)Xu, Hu, Leskovec, and Jegelka]{xu2018how}
K.~Xu, W.~Hu, J.~Leskovec, and S.~Jegelka.
\newblock How powerful are graph neural networks?
\newblock In \emph{ICLR}, 2019.

\bibitem[Ying et~al.(2018)Ying, He, Chen, Eksombatchai, Hamilton, and
  Leskovec]{ying2018graph}
R.~Ying, R.~He, K.~Chen, P.~Eksombatchai, W.~L. Hamilton, and J.~Leskovec.
\newblock Graph convolutional neural networks for web-scale recommender
  systems.
\newblock In \emph{Proceedings of the 24th ACM SIGKDD international conference
  on knowledge discovery \& data mining}, 2018.

\end{thebibliography}
